\documentclass[10pt]{article} 

\usepackage[preprint]{rlj} 

\usepackage{amssymb}            
\usepackage{mathtools}          
\usepackage{mathrsfs}           
\usepackage{graphicx}           
\usepackage{subcaption}         
\usepackage[space]{grffile}     
\usepackage{url}                
\usepackage{lipsum}             

\title{Toward Enactive Artificial Intelligence}

\setrunningtitle{Toward Enactive Artificial Intelligence}

\author{\normalsize Banafsheh Rafiee\textsuperscript{1}, Richard S. Sutton\textsuperscript{2,3}}

\emails{\footnotesize rafiee@ualberta.ca, rsutton@ualberta.ca}
\affiliations{
\small
$^{1}$ Independent Researcher\\
$^{2}$ Department of Computing Science, University of Alberta, Canada\\
$^{3}$ Alberta Machine Intelligence Institute (Amii)\\
}

\contribution{
    Provide a succinct but precise list of the contribution(s) of the paper. Use contextual notes to avoid implications of contributions more significant than intended and to clarify and situate the contribution relative to prior work (see the examples below). If there is no additional context, enter ``None''. Try to keep each contribution to a single sentence, although multiple sentences are allowed when necessary. If using complete sentences, include punctuation. If using a single sentence fragment, you may omit the concluding period. A single contribution can be sufficient, and there is no limit on the number of contributions. Submissions will be judged mostly on the contributions claimed on their cover pages and the evidence provided to support them. Major contributions should not be claimed in the main text if they do not appear on the cover page. Overclaiming can lead to a submission being rejected, so it is important to have well-scoped contribution statements on the cover page.
    }
    {
    None
    }

\contribution{
    The submission template for submissions to RLJ/RLC 2026
    }
    {
    Built from previous RLC/RLJ, ICLR, and TMLR submission templates
    }

\contribution{
    \textit{{[}Example of one contribution and corresponding contextual note for the paper ``Policy gradient methods for reinforcement learning with function approximation'' \citep{Sutton2000}.]}\\ This paper presents an expression for the policy gradient when using function approximation to represent the action-value function.
    }
    {
    Prior work established expressions for the policy gradient without function approximation \citep{Williams1992}.
    }


\summary{The summary appears on the cover page. Although it can be identical to the abstract, it does not have to be. One might choose to omit the stated contributions in the Summary, given that they will be stated in the box below. The original abstract may also be extended to two paragraphs. The authors should ensure that the contents of the cover page fit entirely on a single page. The cover page does \textbf{not} count towards the 8--12 page limit.

\lipsum[1]
}

\begin{document}

\maketitle  

\begin{abstract}
In this paper, we advocate for incorporating enactive approaches to perception and cognition into artificial intelligence (AI).
Enactive approaches view perception as an active, skillful engagement with the world, where agents perceive by acting and by understanding how their actions shape their experience.
This contrasts with classical views that treat perception as a passive internal process in which the brain receives sensory input, processes it, and issues commands for action.
Enactive views emphasize the dynamic, embodied, and interactive character of perception, grounded in the lived experience of agents embedded in their environments.
We identify and develop four key enactive concepts that we find most relevant to AI: experience, action–perception inseparability, autonomy, and embodiment.
Much of mainstream AI, from classical rule-based systems to large language models, has largely neglected these insights, treating cognition as internal processing detached from embodied interaction and intrinsic normativity.
Reinforcement learning (RL), however, exhibits structural resonance with enactive principles through its emphasis on action, agent–environment interaction, feedback-driven adaptation, and agent-centered evaluation.
However, this resonance should not be taken as theoretical equivalence, as RL approximates some enactive insights,but key elements remain absent or weakly developed.
Building on this analysis, we suggest a broader incorporation of enactive ideas into both mainstream AI and RL.
\end{abstract}

\section{Enactive Cognition}
Enactive perspectives on cognition and perception view cognition, perception, and action as mutually constitutive \citep{varela1991embodied}. Perception does not precede action nor does it merely guide it; rather, they unfold together in ongoing interaction with the environment. Actions shape perceptual experience by modulating sensory input, while perception, in turn, is realized through patterns of action. In other words, perception consists of mastering the ways in which our bodily movements produce changes in our sensory stimulations, known as \textit{sensorimotor contingencies} \citep{noe2004action}. To perceive is thus to skillfully navigate these contingencies, to understand how the world reveals itself through active exploration, rather than to passively register incoming data.

Enactive perspectives contrast with classical representationalist views, in which perception is framed as a passive process where sensory input is transmitted through the body’s sensory apparatus and transformed into internal representations of the world.
These internal representations function as stand-ins for the environment, providing an internal map that allows the system to reason about and act upon the environment without requiring the environment's direct presence \citep{Marr1982, Fodor1983}.
Enactive perspectives reject this view, emphasizing continual and real-time interaction with the environment.
In the representationalist views, the brain functions as a central processing unit: it encodes sensory information, manipulates symbols, and generates action plans based on its internal representations, assuming a sharp distinction between the perceiving agent and the external world, with perception reduced to the accurate construction and manipulation of internal representations.
By contrast, in enactive perspectives, the adequacy of perception is not measured by representational fidelity but by the agent’s capacity for skillful engagement with the environment.

The term enactive was first introduced within Enactivism \citep{varela1991embodied}, a subfield of cognitive science, which frames cognition as \textit{enacted} rather than pre-given.
To enact is to bring forth or constitute a meaningful world through embodied engagement with the environment.
\cite{oregan2001sensorimotor} further argued that perception is not something that happens to an organism but something the organism actively does.
These works contributed to a broader research landscape challenging representationalist accounts of cognition, including embodied and situated approaches.
Embodied cognition \citep{clark1997being} emphasized the role of the body in shaping the mind.
Situated cognition \citep{Hutchins1995,Lave1988} posits that cognitive processes are deeply embedded in the activity, tools, and social interactions rather than arising solely from internal processes.

While we borrow the term enactive from enactivism, the perspective developed in this paper is not restricted to that tradition. 
We bring together a broader set of ideas that appear across philosophy, psychology, and cognitive science.
These approaches are not monolithic and differ in important respects.
However, they share a common orientation: they treat cognition as arising through active, embodied, and situated engagement with the environment.
Our aim is therefore not to provide a historical survey, but to articulate a coherent perspective that synthesizes these related strands.

Enactive ideas trace back to phenomenology, drawing on Husserl, Heidegger, and Merleau-Ponty, who emphasized the inseparability of organism and world, bound together through continual engagement.
Husserl, argued that perception is not a matter of constructing internal representations of the world but of attending to things as they are directly given in lived experience.
Heidegger radicalized this insight with his notion of being-in-the-world, which rejects the picture of a detached subject representing the external world. Instead, he emphasized that human existence is fundamentally situated: we do not first perceive the world and then interpret it, but already are immersed in practical, meaningful contexts that shape understanding from the outset.
Merleau-Ponty emphasized the body not as an object among others, but as the medium through which the world is experienced, with perception depending on our bodily abilities to move and interact.

In psychology, related enactive ideas appear in early traditions.
Gestalt psychology proposed that perceptual experience is shaped by the organism’s own organizing activity, anticipating the emphasis on perception as an active, world-involving process \citep{koffka1935principles}.
\cite{goldstein1939organism} highlighted the inseparability of perception, action, and context, viewing the organism’s behavior as an integrated, adaptive response rather than a sequence of internal computations.
Gibson’s ecological approach treated perception as inherently action-dependent: organisms perceive their environment in terms of what actions they can take, possibilities for action that depend on both the situation and the agent’s own embodied capacities. These possibilities for action, or \textit{affordances}, are not fixed properties of the environment but emerge through interaction with the world \citep{gibson1979ecological}.

With this foundation, we turn to enactive ideas that we consider most pertinent for AI research. We identify and develop four key enactive concepts: experience, action–perception inseparability, autonomy, and embodiment, and examine how different AI approaches align with them.
Our discussion spans two levels of analysis: one from the enactive perspectives, and one at the computational level of AI systems. These levels must be distinguished. The comparisons we draw are structural aiming to identify shared patterns and organizing principles rather than claims of equivalence.

\section{Experience}
Enactive cognition is grounded in experience. 
Central to experience is the continual interaction between agent and environment in which each influences and responds to the other.
Perception is not a one-time act of constructing an internal model of the world that functions as a stand-in for the world.
No internal model, however elaborate, can capture the inexhaustible richness and open-ended variability of the world. 
The world is not a fixed set of features or a static object to be represented; it is a dynamic, evolving field of possibilities that unfolds differently depending on the agent’s activity, context, and engagement history. 
In other words, the world is an ongoing situation that the agent co-constitutes through action \citep{varela1991embodied, noe2004action}. 
Because the world exceeds any finite description, the agent must remain in continual interaction with it, drawing on real-time feedback to adjust its actions, recalibrate its expectations, and refine its understanding. 
In this sense, as Rodney Brooks puts it, “the world is its own best model”: the most reliable, up-to-date, and fine-grained information is always available in the world itself, not in internal surrogates.

In the strong enactive sense, experience is not just continual interaction, but skillful, normative, and embodied. First, experience is skillful: through interaction, agents acquire skills that shape how the world appears to them. 
The environment shows up as a field of affordances, soliciting certain activities while resisting others.
Second, experience is normative rather than neutral. The agent’s actions can succeed or fail, fit well or poorly with the situation, and the agent continually adjusts its behavior in response to this ongoing evaluation.
Third, experience is embodied: it is shaped by the ways in which body enables and constrains possible actions. 
In this section, we focus on the continual interaction aspect of experience; the other aspects are addressed in later sections.

In classic rule-based AI, experience was largely absent. 
In \textit{What Computers Can’t Do}, \cite{dreyfus1992what} argues that rule-based symbolic systems fall short of replicating human intelligence precisely because they lack the experiential basis.
Machine learning introduced hints to the importance of experience by centering learning on data—the byproduct of experience. 
However, it did not fully embrace experience, as most methods, especially supervised learning, depend on data gathered and labelled by humans, as opposed to data collected by the agent itself through direct interaction with its environment.
Moreover, supervised learning treats cognition as a one-time process of learning from fixed datasets rather than as continual engagement and adaptation with the environment.

By contrast, reinforcement learning (RL) placed experience at the core of the learning process, enabling agents to gather their own data by actively interacting with their environment. 
Although RL does not capture the full, skillful, embodied, and normative sense of experience in the enactive sense, it approximates its emphasis on continual interaction through learning from action and feedback.
The importance of allowing AI agents to continually interact with the environment and learn from data gathered by themselves has been well articulated by \cite{silver2025welcome}. The central argument is that data must continually improve alongside the agent’s capabilities, a process made possible only through the agent’s own experience.

A related AI research area is continual learning, which studies how agents can learn from a stream of changing data rather than a fixed dataset. While continual learning does not necessarily emphasize agent-gathered data or sustained interaction with the environment, it aligns with enactive views in emphasizing ongoing adaptation under non-stationarity.
Continual learning has many definitions \citep{abel2023definition,khetarpal2022towards,ring1997child}, closely related research areas \citep{thrun1998lifelong,mitchell2018never}, and concepts such as loss of plasticity and catastrophic forgetting \citep{dohare2024loss,mccloskey1989catastrophic}.
Despite differences, these approaches share the insight that agents must learn continually to remain adaptive in complex, non-stationary environments.
Continual learning becomes particularly important under the \textit{Big World Hypothesis} \citep{javed2024big}, which holds that the world is orders of magnitude larger than the agent and thus is perceived as ever-changing from the agent’s perspective. This assumption echoes the enactive perspectives: because the world exceeds what the agent can internally capture, effective behavior requires continual interaction in which the agent continually updates its understanding and skills.

\section{Action–perception inseparability}
Enactive approaches emphasize the inseparability of action and perception, viewing perception as understanding sensorimotor contingencies, that is how actions produce changes in sensory input.
Visual experience, for example, depends on mastering the patterns of change in visual input as a function of eye, head, and body movements, such as knowing that moving the eyes leftward causes objects in the visual field to shift rightward \citep{noe2004action}.
Similarly, auditory perception relies on sensorimotor contingencies: head movements systematically alter the acoustic input at each ear, and these changes in timing and intensity, provide reliable cues for locating sound sources \citep{vandijk2016}.
Tactile perception likewise depends on sensorimotor contingencies: surface qualities such as texture or shape are revealed through changes in sensory input as the hand moves across them. Fine textures produce distinct vibration patterns depending on movement speed and direction, which are learned through active touch \citep{lederman1987hand}.

Crucially, the emphasis on sensorimotor contingencies does not imply that perception is merely the passive monitoring of these patterns. Rather, the enactive claim is stronger: perception is itself a form of skillful activity \citep{noe2004action}—to perceive is to act. 
Rather than waiting for input to arrive and then processing it internally, agents engage in purposeful movements to reveal the structure of their environment. Agents often act in order to reveal, stabilize, or make sense of perceptual information, for example, by moving the head to resolve ambiguity in a scene. 

Perception, in this sense, is thought of as a feedback loop in which the agent's understanding of a situation and its response to it become increasingly refined over time.
As the agent gets a better grasp of the situation, it becomes able to respond more skillfully and in a more refined manner. 
The situation comes to afford more refined and differentiated responses, presenting not merely general possibilities for action but more precise and nuanced affordances that were not previously accessible to the agent. The agent’s more refined responses, in turn, further refine its understanding by revealing new aspects of the situation.
Thus, refined understanding and response mutually reinforce each other.
This feedback loop was referred to as \textit{the intentional arc} by \cite{MerleauPonty1945}. 

In this feedback loop, the agent aims toward a more coherent understanding of the situation, the \textit{maximal grip} \citep{MerleauPonty1945}. 
In this process, the agent continually adjusts its posture, attention, and movements to bring the situation into sharper focus, both perceptually and motorically. The agent is naturally drawn toward states of greater stability, clarity, and alignment between its body and the environment. 
This is done often in a fluid and unreflective manner: much like leaning in to see a detail more clearly or tilting the head to resolve an ambiguous sound, the agent acts in ways that reduce uncertainty and restore perceptual equilibrium \citep{dreyfus2002intelligence}. These refinements are not guided by explicit reasoning but by a bodily sense of tension when the current state deviates from an optimal relation, and by the felt relief that comes with moving closer to that optimal state.

Action–perception inseparability is largely absent in mainstream AI, where perception is dominantly understood as preceding action. 
This mirrors the representationalist view, where perception is treated as passive extraction of information from sensory input, later used to guide behavior.
This view is reinforced by large-scale video-generation systems, where purely observational learning is sometimes taken to yield understanding of phenomena like intuitive physics without active interaction.
Yet the competence of such systems extends only as far as the world continues to behave like their data. 
As noted by \cite{goddu2024llms}, a system that has learned traffic light regularities may accurately predict the sequence of green, yellow, and red, but this amounts only to tracking regularities, not to understanding how the sequence could be changed.
When the light malfunctions, is interrupted, or action is required to alter the situation, such as stopping traffic, triggering a crossing signal, or diagnosing a fault, the system has nothing to fall back on. 
By contrast, an enactive system does not merely anticipate what typically occurs, but can intervene, correct, and explore when expectations fail. The difference is not one of accuracy but of kind: a generative video model can continue a pattern, whereas an enactive system can determine what to do next when the pattern breaks.

In contrast to the mainstream view, several strands of AI have explored action-perception coupling. 
Early work, such as the Pengi system \citep{agre1987pengi}, showed that competent behavior can arise from tightly coupled perception–action loops without explicit internal world models, where perception is directly implicated in triggering action rather than serving as a detached input stage. 
\cite{ballard1991animate} and \cite{whitehead1990active} further demonstrated that perception is an active process of information acquisition: by moving the eyes or body to gather task-relevant information, the agent learns control policies that determine what is sensed, making vision dependent on action strategies rather than passive encoding. Situated planning \citep{chapman1989situated} highlighted that even planning is not separable from perception in realistic settings: plans must be continually revised in response to perceptual feedback, since viable actions depend on the evolving interaction with the environment rather than a pre-given model of it. Behavior-based robotics \citep{brooks1991intelligence} emphasized this coupling at the architectural level, showing that coherent behavior can emerge without centralized representations precisely because perception and action are distributed across interacting sensorimotor routines, rather than separated into sequential stages.
In parallel, formal accounts such as predictive coding, the perception–action cycle, and active inference \citep{rao1999predictive, tishby2010information, friston2017active} framed behavior as a unified loop in which perception and action are jointly organized, either through hierarchical prediction error minimization, information-theoretic optimization, or variational inference under a shared objective.
Complementing these approaches, several formal frameworks capture aspects of action–perception coupling by enabling agents to anticipate how sensory inputs evolve under their actions, and by treating action, observation, and internal state as mutually dependent. This includes General Value Functions \citep{sutton2011horde}, Predictive State Representations \citep{littman2002psr}, Partially Observable Markov Decision Processes \citep{kaelbling1998planning}, and world models \citep{ha2018world}.

Similar ideas appear in recent work. 
\cite{sutton2022reward} introduced the \textit{STOMP} framework in which the agent learns subtasks that maximize distinct aspects of perception. 
As the agent improves at each subtask, it also refines its model of how actions affect perception, creating a feedback cycle in which refined actions lead to refined perception, and vice versa, echoing the \textit{intentional arc}.
\cite{machado2023temporal} proposed a loop where an agent's representation of its environment and behavior are refined through constant interaction. As the agent develops better ways of representing the world, it discovers behaviors, which in turn enrich representations.
Another line develops affordance theory in RL, where the environment constrains and suggests possible actions \citep{graves2022affordance,khetarpal2020can}. 
Rather than treating all actions as always available, agents learn which are afforded in each situation, mirroring selective, skillful responses in enactive perception.

\section{Autonomy}
Autonomy plays a central role in enactive cognition.
Agents are not passive responders to external stimuli, but self-organizing systems whose perception is shaped by their goals and needs \citep{diPaolo2005}.
This autonomy is often grounded in the notion of \textit{autopoiesis}: agents are self-producing and self-maintaining systems that actively sustain their own organization.
As such, perception reflects what matters to the organism from the standpoint of its continued viability \citep{jonas1966, diPaolo2017}. What is perceived is not simply what is present, but what is relevant to the agent’s self-maintenance.  The world does not show up as a neutral array of features, but as meaningful relative to the agent’s goals and needs \citep{gallagher2017}.

Autonomy, in this sense, gives rise to normativity. Because the agent must continually maintain itself as a coherent system, its interactions with the environment are not neutral; they can succeed or fail relative to the agent’s continued viability. Normativity, in this sense, arises from the agent’s need to maintain its own organization, rather than being imposed from the outside.
This normative structure, in turn, shapes what is relevant to the agent. The environment is therefore not encountered merely in terms of what is, but in terms of what matters: what supports or threatens the agent’s ongoing self-maintenance. In other words, it serves to constrain the vast space of possible generalizations: in a world where everything could resemble everything else in countless ways, the agent’s goals and needs filter which similarities are relevant \citep{dreyfus2002intelligence}. What counts as similar is shaped not by abstract resemblance, but by the agent’s ongoing movement toward success in a particular context.

To analyze the notion of autonomy and normativity within AI, it is useful to distinguish between two related questions. First, whether the agent can evaluate its own behavior, that is, whether it has a sense of success and failure grounded in its own activity. Second, whether the criteria for success and failure arise from the agent itself or are imposed externally.

In much of AI, particularly supervised learning, both aspects are absent. The system does not evaluate its own performance; correctness is determined only by comparison to human-provided labels. For any given input, the model has no access to whether its output is successful unless an external signal is supplied. Evaluation is therefore not part of the agent’s ongoing activity, but occurs entirely externally. 
Moreover, success criteria are fully externally specified by the dataset and labeling process. 
Thus, the system neither self-evaluates nor has its own standards of success.
This holds for large language models as well: although trained with self-supervised objectives such as next-token prediction, they effectively learn by imitating patterns in human-generated data and are not able to evaluate their own outputs without outsourcing evaluation to external signals.

A similar pattern appears in classical symbolic planning and automated reasoning \citep{newell1956logic, ghallab2004automated} where success is defined by satisfaction of explicitly specified conditions. This introduces a notion of success and failure, but not as grounded in the agent’s activity. The system does not assess its performance as it acts; instead, evaluation is limited to checking whether predefined conditions are met, typically as a binary outcome. 
There is no ongoing, graded assessment of performance during interaction. Furthermore, success criteria are externally imposed: goals are specified in advance and function as fixed conditions to be achieved.

In planning and control-based systems, evaluation becomes continuous and embedded in the agent’s interaction with the environment. These systems track their state relative to a desired setpoint, trajectory, or cost function, and adjust their behavior through feedback. This enables the system to register deviations and to distinguish, at each moment, whether its behavior is improving relative to the target. In this sense, the system possesses an ongoing, graded assessment of success. However, this evaluation remains tied to instantaneous deviation from a predefined objective, that is, how well the current state matches the target. 
Moreover, success criteria, such as the target state or cost function, are still specified externally. 

RL marks a significant shift in that the agent evaluates its behavior through experience. 
By interacting with the environment and receiving rewards, the agent can assess how well it is doing across entire trajectories, rather than only in terms of its current state.
This allows actions to be evaluated in terms of their consequences, including delayed effects, introducing a temporally extended notion of success. 
While control systems answer "how close am I to the target now?", RL answers "was this behavior good, given what it led to over time?".
In this sense, RL captures an important aspect of normativity: evaluation is embedded in the agent's ongoing interaction and learning process. 
However, evaluation criteria remain externally defined through the reward function.

Several lines of work have sought to relax dependence on externally specified criteria by developing more agent-centered notions of evaluation. 
The perception–action cycle \citep{tishby2010information} and active inference \citep{friston2017active} both make evaluation emergent from the agent’s ongoing interaction with the environment, arising either through optimization of information flow or through minimization of expected surprise.
Intrinsic motivation develops internal reward signals that drive exploration and skill acquisition without external supervision, often based on learning progress, rewarding improved ability to predict or control aspects of the environment \citep{chentanez2004intrinsically, oudeyer2007intrinsic, csimcsek2006intrinsic}. 
Goal discovery approaches \citep{andrychowicz2017hindsight} construct or reinterpret objectives from experience, expanding what counts as achievement beyond external rewards. 
These methods move toward more agent-centered forms of evaluation, in which the standards for success are less rigidly fixed in advance. Nevertheless, full autonomy in the enactive sense, where normativity arises from the agent’s own organization, remains unrealized.

\section{Embodiment}
Enactive perception is fundamentally embodied: the shape, structure, and capacities of the body influence how an organism perceives \citep{clark1997being, Thompson2007}. 
More specifically, perception is the mastery of sensorimotor contingencies that depend on the body. 
Sensorimotor contingencies are not abstract mappings between inputs and outputs that could be specified independently of embodiment; they are grounded in the specific capacities of a body: what movements it can perform, how it can explore, and how its sensory systems are organized. 
Morphological factors such as joint structure, muscle distribution, and sensory placement constrain and enable the forms of engagement available to the agent, and determine the space of possible sensorimotor contingencies. 
Embodied robotics supports this claim: altering a robot’s physical form can significantly change its perceptual abilities \citep{pfeifer2006understanding}. 
The body is therefore not an optional component added after the fact, but the condition for perception to be possible in the first place \citep{MerleauPonty1945}

Embodiment structures what counts as perceptually relevant. The body constrains the forms of engagement available to an agent, and therefore, shapes which similarities in the environment stand out as meaningful. Gibson’s notion of affordances captures this idea: features of the environment appear as “graspable,” “climbable,” or “passable” only relative to the agent’s bodily capacities. Without a body capable of such actions, these distinctions do not exist as meaningful features. In this way, the body does not simply constrain what can be perceived, but provides a structure for understanding the world in terms of action-relevant similarities \citep{dreyfus2002intelligence}.

Embodiment also connects to autonomy through autopoiesis: an agent is autonomous insofar as it is a self-maintaining system that continually produces and preserves its own organization. The body is not peripheral in this process; it is the site where this self-production is realized. The boundaries, processes, and interactions constituting the agent are grounded in its embodied organization. In this sense, embodiment makes autonomy possible: without a body that sustains and regulates itself through interaction with the environment, there is no autonomous system in the enactive sense.

Mainstream AI largely treats perception in a disembodied manner. Even when trained on large-scale multimodal data, many systems learn mappings from inputs to internal representations without any dependence on sensorimotor engagement or bodily structure. In this setting, perception is decoupled from the kinds of structured sensorimotor contingencies that characterize embodied agents, and instead reduced to pattern recognition over static datasets \citep{bender2021dangers, bommasani2021opportunities}. Threfore, these systems remain limited in their ability to develop genuinely situated perception or to adapt through active exploration of novel environments \citep{harnad1990symbol, chemero2009radical}.

Embodied RL and robotics often treat embodiment as an external constraint rather than a constitutive principle of cognition. Many robotic systems still rely on modular architectures that separate perception, planning, and control, preserving a classical decomposition of cognitive functions \citep{brooks1991intelligence, clark1997being}. In such systems, the body functions primarily as an interface for executing precomputed policies, rather than as a source of structure that shapes perception. Similarly, extensive reliance on simulation and offline training further distances learning from the full variability and constraint structure of real sensorimotor interaction \citep{jakobi1995noise, cully2015robots}.

By contrast, work in soft robotics and morphological computation demonstrates that bodily structure can play an active computational role, simplifying control and shaping behavior through physical dynamics \citep{rus2015design, pfeifer2006understanding, zahedi2013quantifying}. Despite these insights, such approaches remain peripheral within mainstream robotics and AI, where disembodied data-driven learning and modular system design continue to dominate. Consequently, the enactive emphasis on embodied sensorimotor engagement has yet to be fully integrated into AI.

\section{Conclusion}
Mainstream AI has largely failed to appreciate the enactive insights. 
By contrast, RL exhibits several structural resonances: It enables agents to generate their own experience through trial-and-error, places action at the center of learning, and introduces a temporally extended notion of evaluation through reward.
However, this alignment is partial. 
Evaluation remains externally specified through reward functions, and is not grounded in the agent’s own organization; action–perception inseparability is not fully realized, with perception still typically treated as preceding action; and embodiment is treated as an implementation detail rather than a constitutive condition for cognition.

We suggest a deeper incorporation of enactive ideas into RL and AI. This paper takes a step toward articulating key enactive concepts and relating them to existing AI frameworks, but it does not operationalize them. A key future direction is to make these ideas more precise and testable. This includes questions such as: 
what constitutes a higher degree of action–perception inseparability? 
What benchmarks capture skillful engagement rather than pattern reproduction? 
What does self-maintenance mean for artificial agents: battery state, hardware integrity, or learned competence?
What counts as embodiment in AI: a robot body or a software agent with tools and APIs?
Addressing such questions is essential for moving from conceptual alignment toward enactive AI in practice.
\bibliography{main}
\bibliographystyle{apalike}


\end{document}